\documentclass[11pt,a4paper]{article}
\usepackage[hyperref]{emnlp2018}
\usepackage{times}
\usepackage{latexsym}
\usepackage{graphicx}
\usepackage{url}
\usepackage{multirow}
\usepackage{subcaption}
\usepackage{makecell}
\aclfinalcopy 

\title{How agents see things:\\On visual representations in an emergent language game}

\author{Diane Bouchacourt \\
Facebook AI Research\\
  {\tt dianeb@fb.com} \\\And
  Marco Baroni \\
Facebook AI Research\\
  {\tt mbaroni@fb.com} \\}

\date{}

\begin{document}
\maketitle
\begin{abstract}
  There is growing interest in the language developed by agents
  interacting in emergent-communication settings. Earlier
  studies have focused on the agents' symbol usage, rather than on
  their representation of visual input. In this paper, we consider the
   referential games of \citet{Lazaridou:etal:2017},
  and investigate the representations the agents develop during their
  evolving interaction. We find that the agents establish successful
  communication by inducing visual representations that almost
  perfectly align with each other, but, surprisingly, do not capture
  the conceptual properties of the objects depicted in the input
  images. We conclude that, if we are interested in developing language-like
  communication systems, we must pay more
  attention to the visual semantics agents associate to the symbols they use.
\end{abstract}

\section{Introduction}
\label{sec:introduction}

There has recently been a revival of interests in language emergence
simulations involving agents interacting in visually-grounded
games. Unlike earlier work
\citep[e.g.,][]{Briscoe:2002,Cangelosi:Parisi:2002,Steels:2012}, many
recent simulations consider realistic visual input, for example, by
playing referential games with real-life pictures
\citep[e.g.,][]{Jorge:etal:2016,Lazaridou:etal:2017,Havrylov:Titov:2017,Lee:etal:2018b,Evtimova:etal:2018}. This
setup allows us to address the exciting issue of whether the needs of
goal-directed communication will lead agents to associate
visually-grounded conceptual representations to discrete symbols,
developing natural-language-like word meanings.  However, while most
studies present some analysis of the agents' symbol usage, they pay
little or no attention to the representation of the visual input that
the agents develop as part of their evolving interaction.

We study here agent representations following the model and setup of
\citet{Lazaridou:etal:2017}. This is an ideal starting point, since it
involves an extremely simple signaling game \citep{Lewis:1969}, that is however played
with naturalistic images, thus allowing us to focus on the question of
how the agents represent these images, and whether such
representations meet our expectations for natural word meanings. In
their first game, Lazaridou's Sender and Receiver are exposed
to the same pair of images, one of them being randomly marked as the
``target''. The Sender always sees the target in the left position,
and it must pick one discrete symbol from a fixed vocabulary to send
to the Receiver.  The Receiver sees the images in random order,
together with the sent symbol, and it tries to guess which image is
the target. In case of success, both players get a payoff of 1. Since
an analysis of vocabulary usage brings inconclusive evidence that the
agents are using the symbols to represent natural concepts (such as
\emph{beaver} or \emph{bayonet}), Lazaridou and colleagues next modify
the game, by presenting to the Sender and the Receiver different
images for each of the two concepts (e.g., the Sender must now signal
that the target is a \emph{beaver}, while seeing a different beaver
from the one shown to the Receiver). This setup should encourage
concept-level thinking, since the two agents should not be able to
communicate about low-level perceptual characteristics of images they
do not share. Lazaridou and colleagues present preliminary evidence
suggesting that, indeed, agents are now developing conceptual symbol
meanings. We replicate Lazaridou's games, and we find that, in both,
the agents develop successfully aligned representations that, however,
are not capturing conceptual properties at all. In what is perhaps our
most striking result, agents trained in either version of the game
succeed at communicating about pseudo-images generated from random
noise (Fig.~\ref{fig:noise}). We conclude that, if we want interactive
agents to develop a vocabulary of words denoting natural meanings,
more attention must be paid to the way in which they are representing
their perceptual input.

\section{Experimental setup}
\label{section:setup}

\paragraph{Architecture} We re-implement Lazaridou's Sender and
Receiver architectures (using their better-behaved ``informed''
Sender). Both agents are feed-forward networks. The \textbf{Sender}
takes image representations as input, it projects them into its own
representational space, compares them, and finally outputs a
probability distribution over vocabulary symbols, from which a single
discrete symbol is then sampled. We report here results obtained with
an output vocabulary of $100$ symbols, but the same patterns were
observed using a range of sizes from $2$ to $1,000$. The
\textbf{Receiver} takes as input the target and distractor input image
representations in random order, as well as the symbol produced by the
sender (as a vocabulary-sized one-hot vector).  It embeds the images
and the symbol into its own representational space, where it performs
a symbol-to-image comparison, producing a probability distribution
over the two images, one of which is selected by sampling from this
distribution. If the Receiver selected the target image, a reward of 1
is assigned to both agents. The whole architecture is jointly trained
by letting the agents play, and updating their parameters with
Reinforce \citep{Williams:1992}. See \citet{Lazaridou:etal:2017} for details. 

\paragraph{Data} Following \citet{Lazaridou:etal:2017}, for each of
the $463$ concepts they used, we randomly sample $100$ images from
ImageNet \citep{Deng:etal:2009}. We construct $50,000$ mini-batches of
$32$ image pairs during training and $1,024$ pairs for validation. We
construct a held-out test set in the same way by sampling $10$ images
per concept from ImageNet (for 2 concepts, we were not able to
assemble enough further images), for a total of $4,610$. We compute
RSA scores (see below) on the cross-product of these images. We also
use the held-out set to construct mini-batches of images pairs to
compute test  performance. Following Lazaridou, the images are
passed through a pre-trained VGG ConvNet
\citep{Simonyan:Zisserman:2015}. The input vector fed to the agents is
the second-to-last $4096$-D fully connected layer\footnote{We found
  very similar results with the top $1000$-D softmax layer.}.

\paragraph{Games} We re-implement both Lazaridou's \textbf{same-image} game,
where Sender and Receiver are shown the same two images (always of
different concepts), and their \textbf{different-image} game, where the Receiver sees
different images than the Sender's. We repeat all
experiments using $100$ random initialization seeds. As we faithfully
reproduced the setup of \citet{Lazaridou:etal:2017}, we refer the
reader there for hyper-parameters and training details.

\section{Experiments}
\label{sec:experiments}

We first asked in which way playing the game affects the way agents
``see'' the input data, that is, in which way their image embeddings
differ from the input image representations, and from each
other. Concerning Sender and Receiver, a reasonable expectation is
that successful communication implies a convergence of
representations. How should these representations relate to the input?
Recall that input representations are from one of the top layers of a
state-of-the-art ConvNet trained on ImageNet concept categorization,
and the top layers of such networks are known to capture high-level
concept semantics \citep{Zeiler:Fergus:2014}. The game image pairs are
always sampled from different concepts. So, it would make sense for
the agents to simply learn to carry through the similarity structure
of the input space, in order to communicate about distinct
concepts. Consequently, we predicted that, as training
proceeds, Sender and Receiver representations will become closer to
each other, and to the input ones.

In order to compare the similarity structure of input, Sender and
Receiver spaces, we borrow \emph{representational similarity analysis}
(RSA) from computational neuroscience
\citep{Kriegeskorte:etal:2008}. Given two sets $r_1$ and $r_2$ of
representations of the same item collection (e.g., $r_1$ is the
collection of input images mapped in Sender embedding space and $r_2$
is the same collection represented by Receiver), we first compute
$s_1$ as all possible pairwise (cosine) similarities between the
representations in $r_1$, and $s_2$ as those in $r_2$. We then compute
the (Spearman) correlation between the similarity vectors $s_1$ and
$s_2$. This latter value, which we will call \emph{RSA score},
measures the global agreement between $s_1$ and $s_2$, relative to the
chosen input collection. If $N$ is the number of items in the
collection that we compute representations for, both similarity
vectors $s_1$ and $s_2$ are of length
$N(N-1)$. 
Therefore, it is not necessary for  representations
$r_1$ and $r_2$ to belong to the same space (for example, in our case,
input and agent vectors have different dimensionality). 

\begin{figure}
\includegraphics[width=\columnwidth]{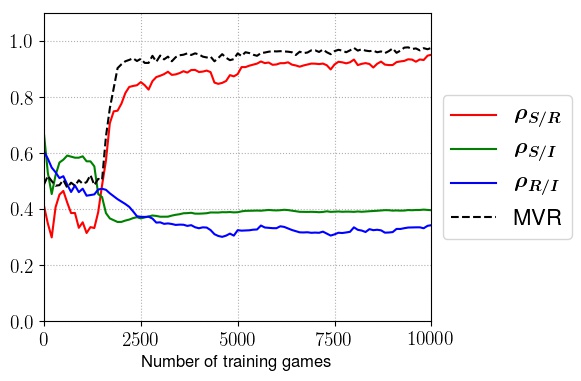}
\caption{RSA scores of the two agents ($\rho_{S/R}$),
and of each agent with the input ($\rho_{S/I}$ and $\rho_{R/I}$), during the
first $10,000$ training games. $S$ refers to Sender, $R$ to Receiver, $I$ to
input. Values remain stable until
end of training. Best viewed in color.}
\label{fig:perf-rsa}
\end{figure}

Figure \ref{fig:perf-rsa} shows RSA and mean validation reward (MVR)
development curves for the cross-validated best seed in the same-image
game. At the beginning of training, the RSA scores are
non-zeros, which is expected as the two agents architectures are
similar and randomly initialized the same way. They are also somewhat
correlated with the input, which we attribute to the fact that
untrained neural networks can already extract relevant image features
\citep{Jarrett:etal:2009}. %
As training converges, Sender and Receiver similarity spaces also
converge.  However, contrary to our prediction, the agent similarity
spaces are not strongly correlated with the input visual space. We
note that, during the first few hundred games, the Sender (green curve)
aligns with the input, but the Receiver (blue curve) does
not. Therefore, it seems that, in order to establish communication,
the two agents have to drift from the input. Indeed, when
communication is successfully established at the end of training,\footnote{We
consider training successful if MVR $\geq{}80\%$.} the
two agents have a RSA score of $\rho_{S/R}=0.98$. However either agent's
score with the input is a much lower
$\rho_{S/I}=\rho_{R/I}=0.33$.\footnote{Values averaged over the $96$
  successful seeds.} On the contrary, when the agents
fail to establish communication, by the end of training their RSA
score is just $\rho_{S/R}=0.39$, but they stay closer  to the
input ($\rho_{S/I}=0.58$ and $\rho_{R/I}=0.42$).\footnote{Values averaged over the $4$
failing seeds.}

The drift of the agents from input similarity could be attributed to
the characteristics of the game they are playing. Since they are only
asked to distinguish between pictures of different concepts, they have
no incentive to keep different instances of a concept distinct (if the
agents are never asked to distinguish one dog from another, they might
eventually become unable to tell dogs apart). That is, we might be
assisting to the inception of a form of categorical perception
\citep{Goldstone:Hendrickson:2009}, whereby the agents lose
sensitivity to within-category differences. If this is the case, we
should observe that same-concept image similarity is \emph{higher} in
Sender (or Receiver) space with respect to input space. However, this
turns out not to be the case. To the contrary, average pairwise
same-concept similarity is consistently \emph{lower} in Sender space
than in the input (mean $z$-normalized same-concept similarity in
input space is at $1.94$ vs.~$0.57$ in Sender space, averaged across
successful seeds). A similar effect is observed by looking at
higher-class (\emph{mammal}, \emph{furniture}, etc.) similarities:
images from the same classes become less similar in Sender space
($0.61$ $z$-normalized within-class input similarity vs.~$0.30$ in
Sender space). This suggests that the agents are becoming less
proficient at capturing the similarity among instances of the same
concept or of the same class. The same conclusion is qualitatively
supported by the pairs of images that underwent the largest
shift between input and Sender space. For example, for two test images of
 avocados which have an input similarity of $0.82$ (and are reasonably similar
 to the human eye), the Sender similarity is at the low value of $-0.27$
 (Receiver similarity is $-0.59$). Contrarily, for an image of a cabin in a
field and an image of a telephone that have an intuitively correct very low input similarity of $0.02$, the Sender similarity for these images is $0.94$ (Receiver similarity is $0.95$).


\citet{Lazaridou:etal:2017} designed their second
game to encourage more general, concept-like referents. %
Unfortunately, we replicate the anomalies above in the different-image
setup, although to a less marked extent. When successful communication
is established at the end of training, the agents have
$\rho_{S/R}=0.90$. But again, the agents' representation do not align
with the input space: their scores with the input are at lower values
of $\rho_ {S/I}=0.40$ and $\rho_{R/I}=0.37$.\footnote{Values averaged
  over the $19$ successful seeds.} In case of communication failure,
by the end of training their RSA score is at the lower value of
$\rho_{S/R}=0.74$, and their values with respect to the input are
$\rho_{S/I}=0.36$ and $\rho_{R/I}=0.34$.\footnote{Values averaged over
  the $81$ failing seeds.} Again, same-concept images drift apart in
agent space, although now to a lesser extent ($1.94$ $z$-normalized
mean similarity in input space vs.~$1.07$ in Sender space). More
encouragingly, we don't find the same pattern for within-class
mean similarities ($0.61$ input space vs.~$0.75$ Sender space).

We must conjecture that the agents are comparing low-level properties
of the image pairs, independently of the game they play. As an extreme
way to test this, we look at how agents trained to play the two games
behave when tested with input pairs that are just random noise
vectors drawn from a standard Normal distribution.\footnote{As during
  training inputs are divided by their norm, we also normalize each
  noise vector, so multiple noise variances would have no
  effect.} If the agents are indeed indifferent to the objects
represented by images, the radical shift in the nature of the input
to the game should not affect them much.

 \begin{table}[tpb]
   \centering
   \begin{footnotesize}
\begin{tabular}{|l|cccc|}
\hline
~ & ~ &\multicolumn{3}{c|}{Test}\\\hline
\multirow{4}{*}{\rotatebox[origin=c]{90}{Train}} & ~ &
Same
im. & Diff.
im. & Noise \\
~ &Same im. &$100$&$72$& $95 $\\
~ & Diff. im. &$98$ &$83$ & $87$\\
\hline
\end{tabular}
   \end{footnotesize}
\caption{Percentage average rewards on the same-image, different-image
  and noise test sets for agents trained in the same- and
  different-image games (chance level at 50\%). For each game,
  values are averaged on $10$ test runs consisting of $1,000$ games of
  mini-batches of $32$ image pairs, using the cross-validated best
  seed.}
\label{tab:tabperf}
\end{table}

\begin{figure}[tbp]
\centering
\begin{subfigure}[t]{0.49\columnwidth}
\centering
\includegraphics[width=\textwidth]{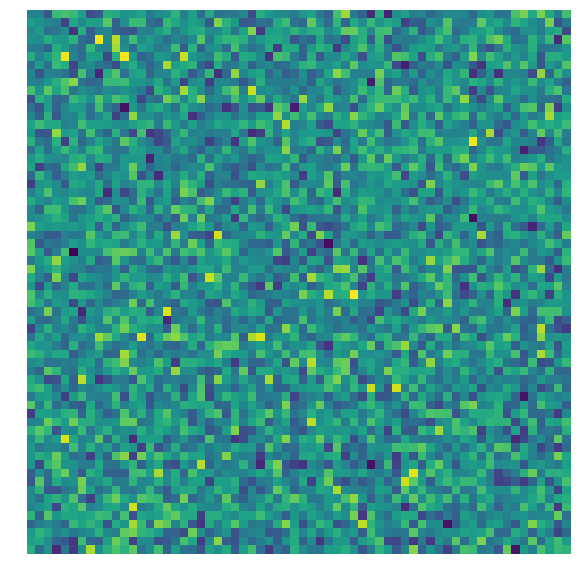}
\end{subfigure}
\hfill
\begin{subfigure}[t]{0.49\columnwidth}
\centering
\includegraphics[width=\textwidth]{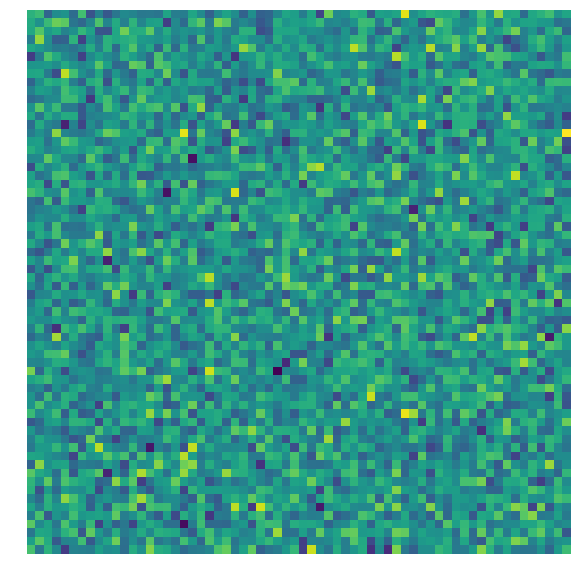}
\end{subfigure}
\caption{Noise vectors agents trained on the
same-image game successfully communicate about.}
\label{fig:noise}
\end{figure}

Results are shown in Table~\ref{tab:tabperf}.  We confirm that the
same-image game is the easiest, and we observe that agents trained in
one game perform reasonably well on the other. More importantly, no
matter which game they are trained on, the agents perform very
well on noise input!  This confirms our hypothesis that the Sender and
Receiver are able to communicate about input data that contain no
conceptual content at all, which in turn suggests that they haven't
extracted any concept-level information (e.g., features that would
allow them to recognize instances of the \emph{dog} or \emph{chair}
category) during training. To get a sense of the sort of noise
pairs agents succeed to communicate about, Figure \ref{fig:noise} provides an example.

Finally, we draw $1,000$ noise pairs $(z_1,z_2)$, and present each to
the Sender with either $z_1$ or $z_2$ as target. We then compare, pair
by pair, whether the highest probability symbol changes when the
target is swapped. We average across $10$ random runs using the best
cross-validated seed. In both versions of the game, for more than
$99\%$ of the pairs, the symbol with highest probability changes when
the target is swapped. This suggests that the agents perform a
\emph{relative} comparison of the two inputs, rather than an absolute
one, in line with the general conclusion that they are not using the
vocabulary to denote stable conceptual properties of the objects
depicted in the images.

\section{Discussion}
\label{sec:discussion}

Existing literature in game theory already showed that convergence towards successful communication is ensured under specific conditions (see ~\citet{Skyrms2010} and references therein). However, the important contribution of \citet{Lazaridou:etal:2017} is to play a signaling game with real-life images instead of artificial symbols. This raises new empirical questions that are not answered by the general mathematical results, such as: When the agents do succeed at communicating, what are the input features they rely upon? Do the internal representations they develop relate to the conceptual properties of the input? Our study suggests that the agents' representations are not capturing general conceptual properties of different objects, but they are rather specifically tuned to successfully distinguish images based on inscrutable low-level relational properties.

Interestingly, our conclusions can be aligned with findings in psycholinguistic experimental literature on dialogue. In order to achieve communication, the agents develop a form of `'conceptual pact'' \citep{Brennan:96}: Their internal representations align while at the same time drifting away from human-level properties of the input. The agents agree on a shared use of the vocabulary, that does not correspond to concepts in the input data.

In future work, we would like to encourage the development of more natural word meanings by enforcing the agent representations to stay more faithful to the
perceptual input they receive. Moving ahead, it is fundamental to design setups where agents would have stronger reasons to develop human-like communication strategies.

\section*{Acknowledgments}
\label{sec:acknowledgments}
We thank Angeliki Lazaridou, Douwe Kiela and Calvin Lee for their
useful discussions and insights. We also thank Francisco Massa for his help on
setting up the experiments.



\end{document}